\documentclass{article}

\PassOptionsToPackage{numbers}{natbib}

\usepackage[preprint]{nips_2018}

\usepackage[utf8]{inputenc} % allow utf-8 input
\usepackage[T1]{fontenc}    % use 8-bit T1 fonts
\usepackage{hyperref}       % hyperlinks
\usepackage{url}            % simple URL typesetting
\usepackage{booktabs}       % professional-quality tables
\usepackage{amsfonts}       % blackboard math symbols
\usepackage{nicefrac}       % compact symbols for 1/2, etc.
\usepackage{microtype}      % microtypography

\usepackage[pdftex]{graphicx}
\usepackage{subcaption}

\title{Uncertainty-Based Out-of-Distribution Detection\\in Deep Reinforcement Learning}

\author{
  Andreas Sedlmeier\thanks{Corresponding author: Andreas Sedlmeier <andreas.sedlmeier@ifi.lmu.de>}\\
  LMU Munich\\
  Munich, Germany\\
  \And
  Thomas Gabor\\
  LMU Munich\\
  Munich, Germany\\
  \AND
  Thomy Phan\\
  LMU Munich\\
  Munich, Germany\\
  \And
  Lenz Belzner\\
  MaibornWolff\\
  Munich, Germany\\
  \And
  Claudia Linnhoff-Popien \\
  LMU Munich \\
  Munich, Germany \\
}

\begin{document}

\maketitle

\begin{abstract}
We consider the problem of detecting out-of-distribution (OOD) samples in deep reinforcement learning.
In a value based reinforcement learning setting,
we propose to use uncertainty estimation techniques directly on the agent's value estimating neural network to detect OOD samples.
The focus of our work lies in analyzing the suitability of approximate Bayesian inference methods and related ensembling techniques that generate uncertainty estimates.
Although prior work has shown that dropout-based variational inference techniques and bootstrap-based approaches can be used to model epistemic uncertainty,
the suitability for detecting OOD samples in deep reinforcement learning remains an open question.
Our results show that uncertainty estimation can be used to differentiate in- from out-of-distribution samples.
Over the complete training process of the reinforcement learning agents,
bootstrap-based approaches tend to produce more reliable epistemic uncertainty estimates, when compared to dropout-based approaches.
\end{abstract}

\section{Introduction}
One of the main impediments to the deployment of machine learning systems in the real world
is the difficulty to show that the system will continue to reliably produce correct predictions in all the situations it encounters in production use.
One of the possible reasons for failure is so called out-of-distribution (OOD) data, i.e. data which deviates substantially from the data encountered during training.
As the fundamental problem of limited training data seems unsolvable for most cases, especially in sequential decision making tasks like reinforcement learning,
a possible first step towards a solution is to detect and report the occurance of OOD data.
This can prevent silent failures caused by wrong predictions of the machine learning system, for example by handing control over to a human supervisor \cite{amodei16}.
In this paper, we propose to use uncertainty estimation techniques in combination with value-based reinforcement learning to detect OOD samples.
We focus on deep Q-Learning \cite{deepQ13}, integrating directly with the agent's value-estimating neural network.

When considering to use uncertainty estimation in order to detect OOD samples, it is important to differentiate two types of uncertainty: aleatoric and epistemic uncertainty.
The first type, aleatoric uncertainty models the inherent stochasticity in the system and consequently cannot be reduced by capturing more data.
Epistemic uncertainty by contrast arises out of a lack of sufficient data to exactly infer the underlying system's data generating function.
Consequently, epistemic uncertainty tends to be higher in areas of low data density.
Qazaz \cite{qazaz96}, who in turn refers to Bishop \cite{bishop94} for the initial conjecture, 
showed that the epistemic uncertainty $\sigma_{epis}(x)$ is approximately inversely proportional to the density $p(x)$ of the input data,
for the case of generalized linear regression models as well as multi-layer neural networks:
$
\sigma_{epis}(x) \propto p^{-1}(x)
$

This also forms the basis of our proposed method: to use this inverse relation between epistemic uncertainty and data density in order to differentiate in- from out-of-distribution samples.

\section{Related Work}
A systematic way to deal with uncertainty is via Bayesian inference.
Its combination with neural networks in the form of Bayesian neural networks is realised by placing a probability distribution over the weight-values of the network \cite{mackay92}.
As calculating the exact Bayesian posterior quickly becomes computationally intractable for deep models, a popular solution are approximate inference methods
\cite{graves11,hernandez15,blundell15,gal16dropout,hernandez16,li17,galConcrete17}.
Another option is the construction of model ensembles, e.g., based on the idea of the statistical bootstrap \cite{efron92}.
The resulting distribution of the ensemble predictions can then be used to approximate the uncertainty \cite{bootstrappedDQN16,lakshminarayanan17}.
Both approaches have been used for tasks as diverse as machine vision \cite{whatUncertainties17}, disease detection \cite{leibig17}, or decision making \cite{depeweg16,bootstrappedDQN16}.

For the case of low-dimensional feature spaces, OOD detection (also called novelty detection) is a well-researched problem.
For a survey on the topic, see e.g. \citet{pimentel14}, who distinguish between probabilistic, distance-based, reconstruction-based, domain-based and information theoretic methods.
During the last years, several new methods based on deep neural networks were proposed for high-dimensional cases, mostly focusing on classification tasks, e.g. image classification.
\citet{hendrycks16} propose a baseline for detecting OOD examples in neural networks, based on the predicted class probabilities of a softmax classifier.
\citet{liang17} improve upon this baseline by using temperature scaling and by adding perturbations to the input.
These methods are not directly applicable to our focus, value-based reinforcement learning, where neural networks are used for regression tasks.
Other methods, especially generative-neural-network-based techniques \cite{schlegl17} could provide a solution, but at the cost of adding separate, additional components.
Our approach has the benefit of not needing additional components, as it directly integrates with the neural network used for value estimation.

\section{Experimental Setup}
One of the problems in researching OOD detection for reinforcement learning is the lack of datasets or environments
which can be used for generating and assessing OOD samples in a controlled and reproducible way.
By contrast to the field of image classification, where benchmark datasets like \textit{notMNIST} \cite{notmnist11} exist that contain OOD samples,
there are no equivalent sets for reinforcement learning.
As a first step, we developed a simple gridworld environment, which allows modifications after the training process, thus producing OOD states during evaluation.

For our experiments, we focus on a simple gridworld pathfinding environment.
During training, the agent starts every episode at a random position in the left half of the $12 \times 4$ grid space.
Its goal is to reach a specific target position in the right half of the grid, which also varies randomly every episode, by choosing one of the four possible actions: $\{\textit{up,down,left,right}\}$.
A vertical set of walls separates the two halves of the environment, acting as static obstacles.
Each step of the agent incurs a cost of $-1$ except the target-reaching action, which is rewarded with $+100$ and ends the episode.
This configuration of the environment is called the \textit{train} environment.
For evaluating  the OOD detection performance, we flip the start and goal positions,
i.e. the agent starts in the right half of the environment and has to reach a goal position in the left half.
This so called \textit{mirror} environment produces states which the agent has not encountered during training.
Consequently, we expect higher epistemic uncertainty values for these OOD states.
Note that training is solely performed in the \textit{train} environment.
Evaluation runs are executed independently of the training process,
based on model snapshots generated at the respective training episodes.
Data collected during evaluation runs is not used for training.
The state of the environment is represented as a stack of three $W \times H$ feature planes ($W$ being the width, $H$ the height of the grid layout)
with each plane representing the spatial positions of all environment objects of a specific type: agent, target or wall.

We compare different neural network architectures and their effect on the reported uncertainty values as the networks are being used by the RL agent for value estimation.
The Monte-Carlo Dropout network (MCD) uses dropout variational inference as described by \cite{whatUncertainties17}.
Our implementation consists of two fully-connected hidden layers with 64 neurons each, followed by two separate neurons in the output layer representing $\mu$ and $\sigma$ of a normal distribution.
Before every weight layer in the model, a dropout layer with $p=0.95$ is added, specifying the probability that a neuron stays active.
Model loss is calculated by minimizing the negative log-likelihood of the predicted output distribution.
Epistemic uncertainty as part of the total predictive uncertainty is then calculated according to the following formula:
$
\textrm{Var}_{ep}(y) \approx \frac{1}{T}\sum_{t=1}^{T}{\hat{y}_t^2} - (\frac{1}{T}\sum_{t=1}^{T}\hat{y}_t)^2
$
with $T$ outputs $\hat{y}_t$ of the Monte-Carlo sampling.

\citet{galConcrete17} suggested an improvement to the default Monte-Carlo dropout method called \textit{Concrete Dropout}
which does not require a pre-specified dropout rate and instead learns individual dropout rates per layer.
This method is of special interest when used in the context of reinforcement learning, as here the available data change during the training process,
rendering a manual optimization of the dropout rate hyperparameter even more difficult.
Our implementation of the Monte-Carlo Concrete Dropout network (MCCD) is identical to the MCD network with the exception that every normal dropout layer is replaced by a concrete dropout layer.
For both the MCD and MCCD networks, 10 Monte-Carlo forward passes are performed.

The Bootstrap neural network (BOOT) is based on the architecture described by \cite{bootstrappedDQN16}.
It represents an efficient implementation of the bootstrap principle by sharing a set of hidden layers between all members of the ensemble.
Our implementation consists of two fully-connected hidden layers with 64 neurons each, which are shared between all heads, followed by an output layer of $K=10$ bootstrap heads.
For each datapoint, a Boolean mask of length equal to the number of heads is generated, which determines the heads this datapoint is visible to.
The mask's values are set by drawing $K$ times from a Bernoulli distribution with $p=0.2$.
The Bootstrap-Prior neural network (BOOTP) is based on the extension presented in \cite{osband-prior18}.
It has the same basic architecture as the BOOT network but with the addition of a so-called random \textit{Prior Network}.
Predictions are generated by adding the output of this untrainable prior network to the output of the different bootstrap heads before calculating the loss.
\citet{osband-prior18} conjecture that the addition of this randomized prior function outperforms ensemble-based methods without explicit priors,
as for the latter, the initial weights have to act both as prior and training initializer.

\section{Results}

\begin{figure}
\centering
\begin{subfigure}{.8\textwidth}
  \centering
   \includegraphics[width=0.8\linewidth]{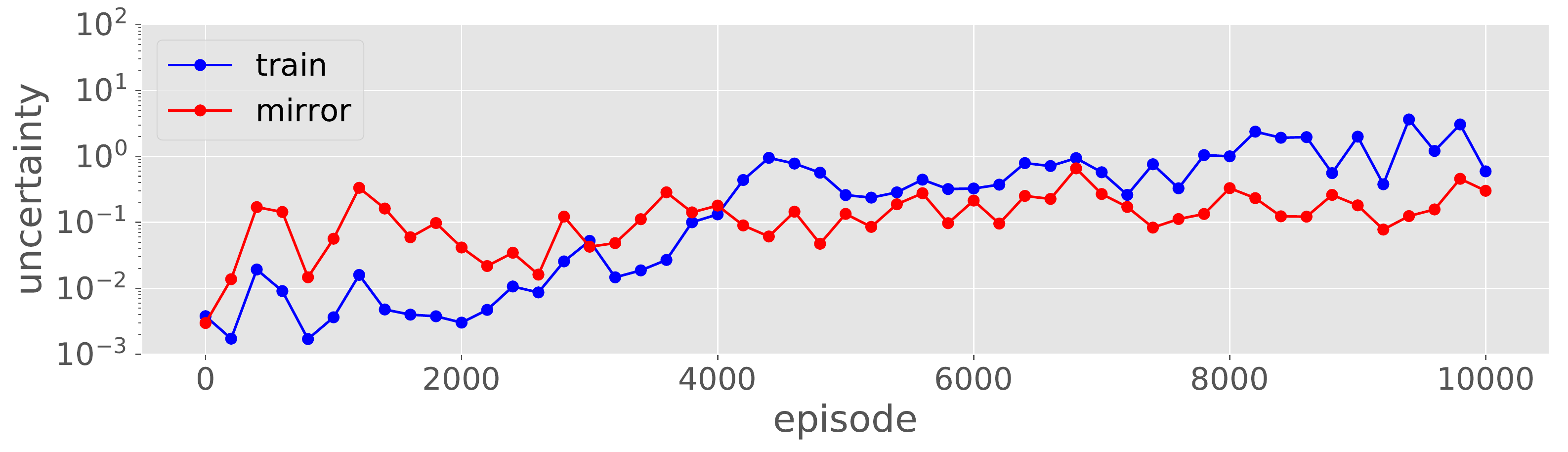}
   \caption{MCD}
   \label{fig:MCD}
\end{subfigure}
\begin{subfigure}{.8\textwidth}
  \centering
   \includegraphics[width=0.8\linewidth]{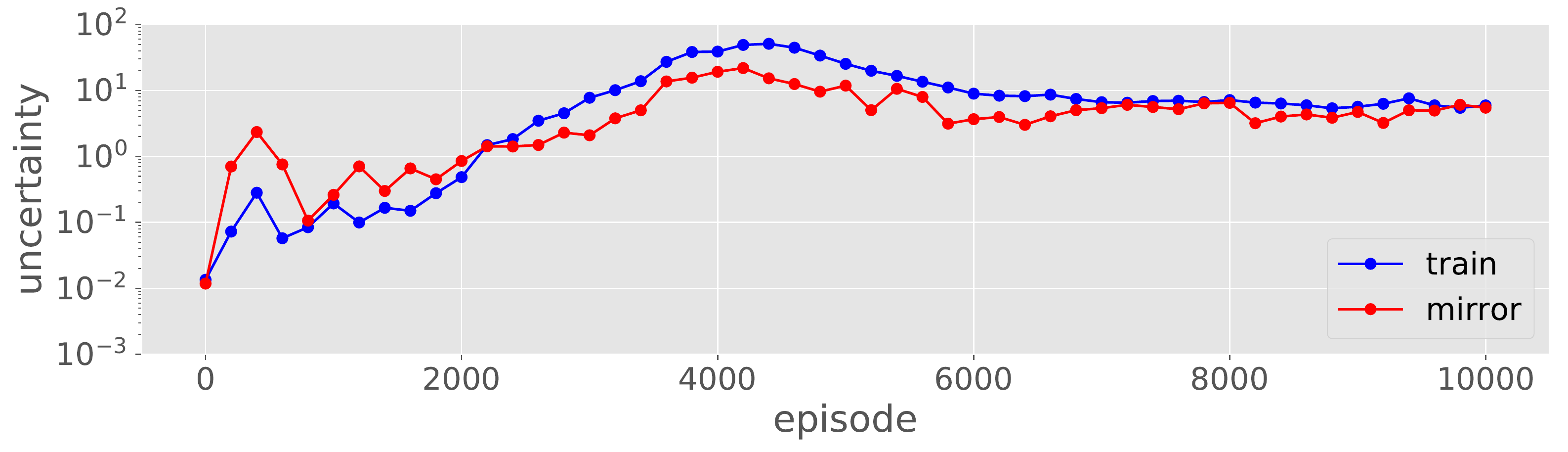}
   \caption{MCCD}
   \label{fig:MCCD}
\end{subfigure}
\begin{subfigure}{.8\textwidth}
  \centering
   \includegraphics[width=0.8\linewidth]{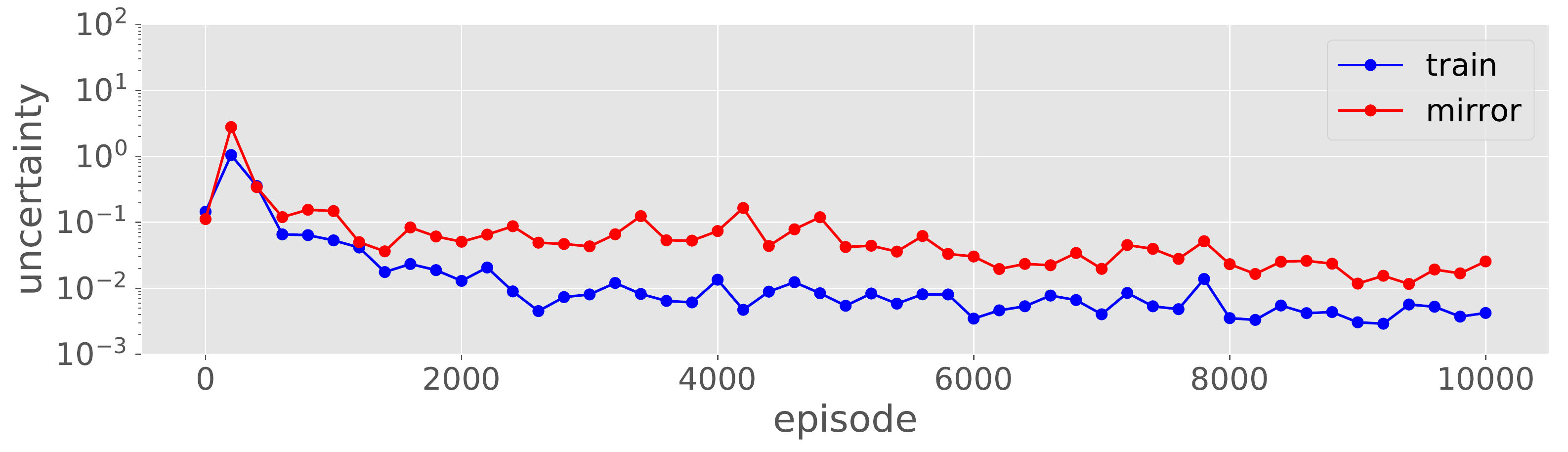}
   \caption{BOOT}
   \label{fig:BOOT}
\end{subfigure}
\begin{subfigure}{.8\textwidth}
  \centering
   \includegraphics[width=0.8\linewidth]{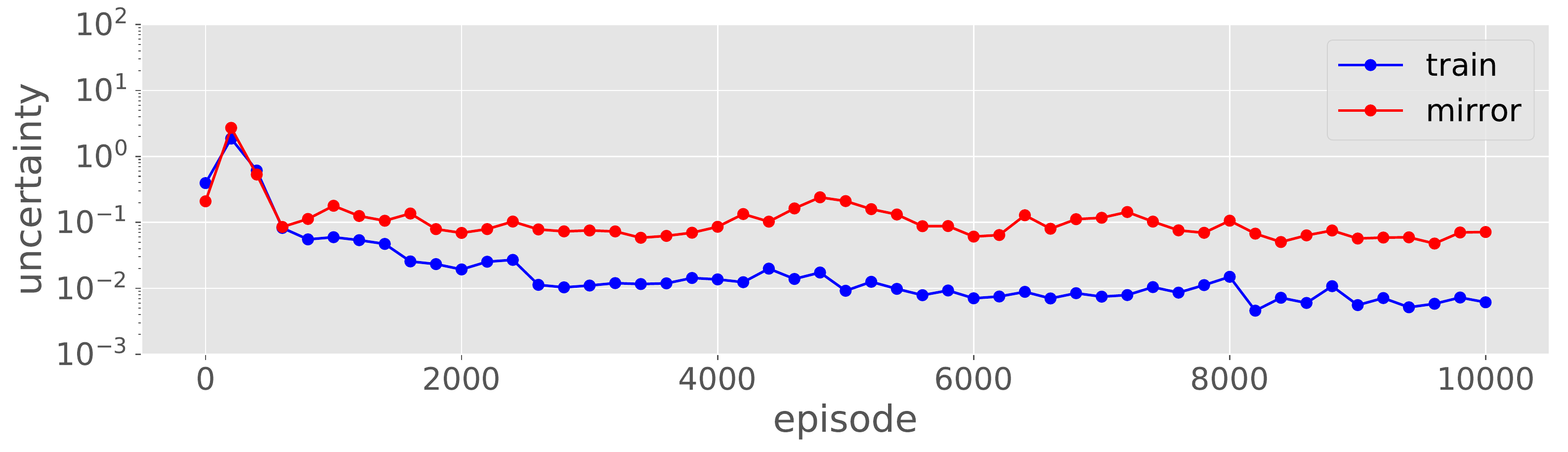}
   \caption{BOOTP}
   \label{fig:BOOTP}
\end{subfigure}
\caption{Per-episode mean uncertainty of chosen actions, averages of 30 runs (y-axis log-scaled).}
\label{fig:uncertainties_models}
\end{figure}

Figure~\ref{fig:uncertainties_models} presents the average uncertainty of the chosen actions over 10000 training episodes of the different network architectures.
As there is a certain amount of randomness in the evaluation runs, caused by the random placement of start and goal positions, the plots show averages of 30 evaluation runs.

According to the concept of epistemic uncertainty, we would expect a decline in the absolute value of reported epistemic uncertainty in the \textit{train} environment over the training process, as the agent collects more data.
Interestingly, only the bootstrap-based methods (BOOT \ref{fig:BOOT} and BOOTP \ref{fig:BOOTP}) reliably show this behaviour.
The dropout-based methods do not show consistent behaviour in this regard.
For these methods, the predicted uncertainty sometimes even increases along the training process as can be seen in Figure~\ref{fig:MCCD}.
Regarding the OOD detection performance, the methods are required to predict higher epistemic uncertainty values for OOD samples than for in-distribution samples.
Here also, the bootstrap-based methods outperform the dropout-based ones.
For all bootstrap methods, over the complete training process, the predicted uncertainty values in the ``out-of-distribution'' \textit{mirror} environment are higher than the values in the \textit{train} environment.
Consequently, it would be possible to detect the OOD samples reliably, for example by setting a threshold based on the lower uncertainty values predicted during training.
Figure~\ref{fig:BOOTP} shows that the addition of a prior has a positive effect on the separation of in- and out-of-distribution samples, as the distance between the predicted uncertainty values increases.

Our results for the dropout-based techniques are not as positive. As can be seen in Figure~\ref{fig:MCD} and \ref{fig:MCCD}, 
neither of the tested Monte-Carlo dropout methods consistenly outputs higher uncertainty values for the OOD states of the \textit{mirror} environment over the complete training process.
Although there are episodes, especially in the beginning, where the \textit{mirror} environment's uncertainty values are higher, there is a reversal during the training process.
As a consequence, it would not be possible to reliably differentiate between in- and out-of-distribution samples at every point in time.

\section{Discussion and Future Work}
The results we obtained from the bootstrap-based methods show the general feasibility of our approach, 
as they allow for a reliable differentiation between in- and out-of-distribution samples in the evaluated environments.
Declining uncertainty values over the training process also conform to the expectation that epistemic uncertainty can be reduced by collecting more data.
For the dropout-based techniques, it remains to be seen if our results show a general problem of these methods in sequential decision problems, 
or whether the results are a consequence of our specific environments.
According to \citet{osband-prior18} the observed behaviour is to be expected for the basic Monte-Carlo dropout method (MCD) as the dropout distribution does not concentrate with observed data.
Consequently, we expected different results from the concrete dropout method (MCCD) as it should be able to adapt to the training data.
Nevertheless, this did not lead to decreasing uncertainty estimates over the training process or a reliable prediction of higher uncertainty for OOD samples.
We are currently working on extending our evaluations to more environments in order to evaluate if these results generalize.
This will include stochastic domains, where it is necessary to differentiate between aleatoric and epistemic uncertainty in order to correctly detect OOD samples.
It will also be very interesting to compare the performance of the proposed uncertainty based methods to methods based on generative models.
Another interesting aspect which could further improve the OOD detection performance of the ensemble methods is the choice of prior \cite{noiseContrastive18}
and a newly proposed method called \textit{Bayesian Ensembling} \cite{bayesianEnsembling18}, which could bridge the gap between fully Bayesian methods and ensembling methods.

\bibliographystyle{plainnat}
\bibliography{ood_rl_nips_2018}

\end{document}